\documentclass[journal ]{IEEEtran}

\usepackage{times}
\usepackage{epsfig}
\usepackage{graphicx}
\usepackage{amsmath}
\usepackage{amssymb}
\usepackage[table]{xcolor}
\usepackage[noend]{algpseudocode}
\usepackage{enumitem} 
\usepackage{cite}
\usepackage{lineno}

\usepackage{multirow}
\newcommand{\argmin}{\operatornamewithlimits{arg\ min}} 
\usepackage[linesnumbered,ruled]{algorithm2e}
\usepackage{ctable} 
\usepackage[pagebackref=true,breaklinks=true,letterpaper=true,colorlinks,bookmarks=false]{hyperref}

\ifCLASSINFOpdf
\else
\fi

\begin{document}
%
\title{Knowledge-Guided Deep Fractal Neural Networks \\ for Human Pose Estimation}
%
%
%

\author{Guanghan Ning,~\IEEEmembership{Student Member,~IEEE,}
        Zhi Zhang,~\IEEEmembership{Student Member,~IEEE,}
        and~Zhihai He,~\IEEEmembership{Fellow,~IEEE}
}

\maketitle

\begin{abstract}
Human pose estimation using deep neural networks aims to map input images with large variations into multiple body keypoints which must satisfy a set of geometric constraints and inter-dependency imposed by the human body model. This is a very challenging nonlinear manifold learning process in a very high dimensional feature space. 
We believe that the deep neural network, which is inherently an algebraic computation system, is not the most effecient way to capture highly sophisticated human knowledge, for example those highly coupled geometric characteristics and interdependence between keypoints in human poses. 
In this work, we propose to explore how external knowledge can be effectively represented and injected into the deep neural networks to guide its training process using learned projections that impose proper prior. Specifically, we use the stacked hourglass design and inception-resnet module to construct a fractal network to regress human pose images into heatmaps with no explicit graphical modeling.
We encode external knowledge with visual features which are able to characterize the constraints of human body models and evaluate the fitness of intermediate network output. We then inject these external features into the neural network using a projection matrix learned using an auxiliary cost function. The effectiveness of the proposed inception-resnet module and the benefit in guided learning with knowledge projection is evaluated on two widely used human pose estimation benchmarks. Our approach achieves state-of-the-art performance on both datasets. 
\end{abstract}

\begin{IEEEkeywords}
Human Pose Estimation, Fractal Networks, Knowledge-Guided Learning.
\end{IEEEkeywords}

%
\IEEEpeerreviewmaketitle

\section{Introduction}
%
%
%
%
\IEEEPARstart{T}{he} task of human pose estimation is to determine the precise pixel locations of body keypoints from a single input image \cite{fu2017orgm, dantone2014body, zhang2014human, jiang2011human, zhao2015learning, li2017human, eichner2012human}. Closely-related tasks include 3D human pose estimation \cite{belagiannis20163d} and human pose estimation in videos \cite{zhou2016spatio, pfister2015flowing}.
Human pose estimation is very important for many high-level computer vision tasks, including action and activity recognition \cite{ikizler2012web, marcos2015let, cai2016effective}, semantic content retrieval \cite{ren2012visual}, human-computer interaction, motion capture \cite{ kadu2014automatic}, and animation.
Estimating human poses from still images is a challenging task. An effective human pose estimation system must be able to handle large pose variations, changes in clothing and lighting conditions, severe body deformations, heavy body occlusions \cite{toshev2014deeppose, tompson2014joint, newell2016stacked}.
A key question for addressing these problems is how to extract strong low and mid-level appearance features capturing discriminative as well as relevant contextual information and how to model complex part relationships allowing for effective yet efficient pose inference.
Traditional methods for pose estimation are mostly based on Pictorial Structure (PS) models \cite{sapp2013modec, pishchulin2013poselet, sun2011articulated, tian2012exploring, dantone2013human, karlinsky2012using}, which models the spatial relations of rigid body parts using a tree model.
A major drawback of such models is the need to hand-design the structure of the model in order to capture important problem-specific dependencies amongst the different output variables and at the same time allow for tractable inference.

\begin{figure}[!t]
\centering
\includegraphics[width=0.8\linewidth]{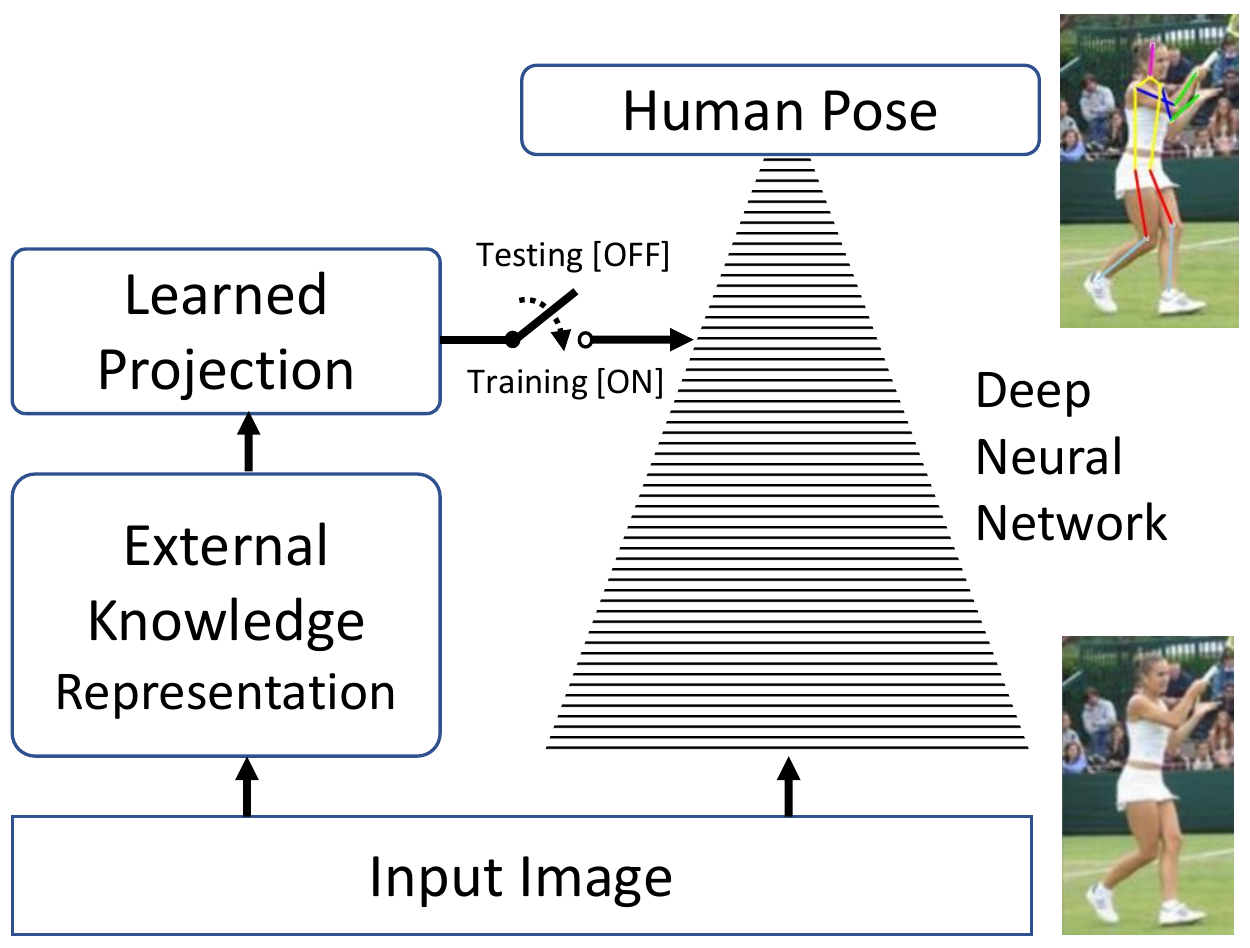}
\caption{\textbf{Knowledge projection for guided learning}. We encode external knowledge visual features which characterizes the constraints of human body models and then inject these external features into the neural network using a projection matrix learned using an auxiliary cost function, which is removed during testing, therefore not increasing network complexity. }
\label{fig:intro}
\end{figure}

With Convolutional Neural Networks (ConvNets) and many assistive methods such as batch normalization \cite{ioffe2015batch}, resnet \cite{he2016deep}, and inception design \cite{szegedy2015going, szegedy2016rethinking}, human pose estimation has recently achieved significant progress.
Even though deep neural networks are capable of fitting large training data through extensive training, the network often needs to be constructed deeper and wider to gain enough representation power \cite{ba2014deep}. As the network becomes more complex, the learning and training processing become more sophisticated and challenging \cite{romero2014fitnets}, especially for those applications with complicated loss functions.

Human pose estimation using deep neural networks requires us to map the input images with large variations into multiple body keypoints which must satisfy a set of geometric constraints and interdependence imposed by the human body model. This is a very challenging nonlinear manifold learning process in a very high dimensional feature space. 
We believe that the deep neural network, which is inherently an algebraic computation system, is not the most efficient way to capture highly sophisticated human knowledge, for example those highly coupled geometric characteristics and interdependence between keypoints in human poses. 

In this work, we propose to explore how external knowledge can be effectively represented and injected into the deep neural networks to guide its training process using learned projections for more accurate and robust human pose estimation. Specifically, as illustrated in Fig. \ref{fig:framework}, we use inception-resnet module and the stacked hourglass structure to construct a fractal network to regress human pose images into heatmaps with no explicit graphical modeling. 
We encode external knowledge with visual features which characterize the constraints of human body models and evaluate the fitness of intermediate network output. We then inject these external features into the neural network using a projection matrix learned using an auxiliary cost function.
The guidance from the external knowledge is only used during the training process, and is turned off during network inference for human pose estimation. 
The benefit of external knowledge is to guide the training of the neural network. Its effect is implicitly imposed on the tuning of the parameters, instead of explicit feature representation of the network.
The injected features for pairs of limbs impose a strong prior during the training, preventing human part keypoint from connecting to noises, e.g., keypoint from other people in the background that is not cropped out for the target person.

The major contributions of this work are summarized as follows:
(1) We develop a new framework to represent and project human knowledge to guide the training of deep neural networks for human pose estimation. This external knowledge project framework is generic and can be extended to other learning and training applications and deep neural network design. 
(2) We propose an efficient network structure, called {\it fractal networks}, for human pose estimation to capture the multi-scale interdependence between body joints in the pose model. This fractal network uses an \textit{inception-resnet module} as the building block.

The rest of the paper is organized as follows.
In section \ref{sec:related-work}, we provide a brief review of recent works on human pose estimation. Section \ref{sec:proposed-method} introduces the concept of knowledge guided learning, the structure of fractal network, and the design of inception-resnet module. Section \ref{sec:experiments} presents our experimental results. Section \ref{sec:conclusions} concludes our paper.

\section{Related Work}
\label{sec:related-work}

\begin{figure*}
	\begin{center}
		\includegraphics[width=1.0\linewidth]{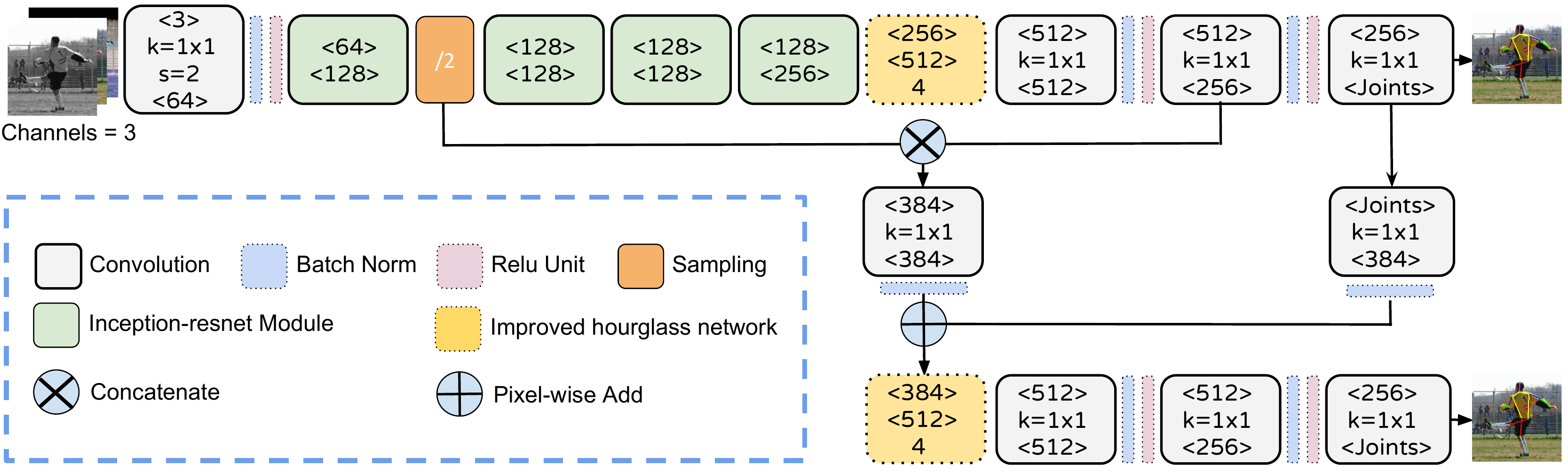}
	\end{center}
	\caption{\textbf{Overview of fractal network}. The network is fractal in that it reflects the concurrence of inception and residual design at both the highest and lowest (inception-resnet module) levels of abstractions. At top level, images of size $256 \times 256$ are down-sampled into the resolution of $64 \times 64$. Subsequently, inputs and outputs of all modules are of size $64 \times 64$, including the output heatmaps. The numbers within brackets in each module denote the number of input and output channels, respectively.}
	\label{fig:overview}
\end{figure*}

\subsection{Structured Prediction and Graphical Models}
Prior to the advent of neural networks most previous work was based on pictorial structures \cite{felzenszwalb2005pictorial} which model the human body as a collection of rigid templates and a set of pairwise potentials taking the form of a tree structure, thus allowing for efficient
and exact inference at test time.
Higher knowledge of the human body is exploited by modeling humans with body parts that are connected via a skeleton structure.
Pictorial structure model \cite{fischler1973representation, felzenszwalb2005pictorial}, models the spatial relations of rigid body parts using a tree model.
A pre-defined kinematic body model is often used to assume that each body part is
independent of all the others except for the ones it is attached to.
A major drawback of such models is the need to hand-design the structure of the model in order to capture important problem-specific dependencies amongst the different output variables and at the same time allow for tractable inference.

Recent work includes sophisticated extensions like mixture, hierarchical, multimodal and strong appearance models \cite{yang2013articulated, pishchulin2013poselet, tian2012exploring, sapp2013modec, pishchulin2013strong}, non-tree models \cite{karlinsky2012using, dantone2013human} as well as cascaded/sequential prediction models like pose machines \cite{ramakrishna2014pose}.
While in \cite{felzenszwalb2005pictorial} each limb is represented by a single template that is parameterized by location, orientation, shape parameters, and an appearance model, Yang and
Ramanan \cite{yang2013articulated} propose mixtures of part templates where body part is represented by
a set of deformable part templates.
Although this approach performs well in comparison to classical pictorial structure
models for human pose estimation, it has some limitations. For instance, the used
scanning-window templates trained with linear SVMs and HOG features \cite{dalal2005histograms} are very
sensitive to noise \cite{zhu2012we}.
Hierarchical models \cite{sun2011articulated, tian2012exploring} represent the relationships between parts at different scales and sizes in a hierarchical tree structure. The underlying assumption of these models is that larger parts (that correspond to full limbs instead of joints) can often have discriminative image structure that can be easier to detect and consequently help reason about the location of smaller, harder-to-detect parts. On the other hand, there are non-tree models \cite{dantone2013human, karlinsky2012using} to incorporate interactions that introduce loops to augment the tree structure with additional edges that capture symmetry, occlusion and long-range relationships. These methods usually have to rely on approximate inference during both learning and at test time.

\subsection{Deep Neural Networks for Human Pose Regression}
ConvNets have been shown to produce remarkable performance for a variety of difficult
Computer Vision tasks including detection \cite{ren2015faster, redmon2016you}, recognition \cite{he2016deep, simonyan2014very}, and semantic segmentation \cite{shelhamer2017fully}.
A key feature of these approaches is that they integrate non-linear hierarchical feature extraction with the classification or regression task in hand being also able to capitalize on
very large data sets that are now readily available.

Since the work of \textit{DeepPose} by Toshev {\it et al.} \cite{toshev2014deeppose}, research on human pose estimation has shifted from traditional approaches to deep neural networks (DNN) due to their superior performance. 
In the context of human pose estimation, it is natural to formulate the problem as a regression one in which CNN features are regressed in order to provide joint predictions of the body parts \cite{toshev2014deeppose, pfister2015flowing, belagiannis2015robust}. For the case of non-visible parts, learning the complex mapping from occluded part appearances to part locations is hard and the network has to rely on contextual information provided by other visible parts to infer the occluded part locations.
DeepPose uses a deep neural network to directly regress the coordinates of body joints. Tompson {\it et al.} \cite{tompson2014joint} argued that it is more efficient to use DNN to regress heatmap images at multiple scales.
While body models are not a necessary component for effective part localization, constraints between parts allow us to assemble independent detections into a body configuration. 
Detection-based methods are relying on powerful CNN-based part detectors which are then combined
using a graphical model \cite{chen2014articulated, tompson2014joint} or refined using regression \cite{tompson2015efficient, pishchulin2016deepcut}. Regression-based methods try to learn a mapping from image and CNN features to part locations. 
\cite{chen2014articulated} achieved promising results by combining CNN-based body part detectors with a body model \cite{yang2013articulated}. 

Human pose estimation methods using deep neural networks have proven their significant advantages over traditional approaches.
However, deeper and wider networks are often required to improve the feature representation power, which in turn leads to increased difficulty in training the neural networks.
Recently, residual learning \cite{he2016deep} has been used to significantly improve the performance of human pose estimation \cite{insafutdinov2016deepercut, newell2016stacked}.
It was used for part detection in the system of \cite{insafutdinov2016deepercut}. \textit{stacked hourglass network} of \cite{newell2016stacked} elegantly extends fully convolutional networks \cite{long2015fully} and deconvolution nets \cite{zeiler2011adaptive} with residual learning.

Intermediate supervision \cite{wang2015training}, recursive prediction \cite{belagiannis2016recurrent}, and inception design \cite{szegedy2015going, szegedy2016rethinking} are among other successful techniques that have been applied by recent methods for human pose estimation. 
Recently, researchers recognize that successive predictions can boost the performance of pose estimation, where parts are sequentially refined \cite{toshev2014deeppose, ramakrishna2014pose, wei2016convolutional, carreira2016human}. In these models an initial prediction is made of all the parts; in subsequent steps, all part predictions are refined based on the image and earlier part predictions.
Tompson {\it et al.} \cite{tompson2015efficient} use a cascade of networks for refined predictions to achieve significantly improved precision in joint localization.
Carreira {\it et al.} \cite{carreira2016human} introduce a so-called \textit{Iterative Error Feedback} scheme, where a set of predictions is included in the input, and each pass through the network further refines these predictions. Their method requires multi-stage training and the weights are shared between iterations.
Recently, adding supervision to intermediate layers of deep networks is also explored to assist the training process \cite{szegedy2015going, lee2015deeply}.
Methods in \cite{newell2016stacked, wei2016convolutional, insafutdinov2016deepercut, belagiannis2016recurrent} use intermediate supervision to add auxiliary supervision branches 
in the network to assist the training process for human pose estimation. These approaches all employ the inception design by concatenating heatmaps from different stages or abstract levels as the input for the next layers.

One direction for further improvement of human pose estimation is to design convolutional networks that can produce robust visual features. Multi-scale processing by repetitive down-sampling and up-sampling has been introduced in \textit{Stacked Hourglass Networks} \cite{newell2016stacked}.
Another approach to improve human pose estimation performance is to use explicit part-based models \cite{karlinsky2012using, dantone2013human, yang2013articulated} or implicitly encode configuration model using its contexts \cite{bulat2016human}. These methods involve additional sub-networks to detect parts, which increases the overall complexity. 
In this work, we leverage these ideas and approaches. We propose a fractal network structure using inception-resnet as building blocks to explore the multi-scale interdependence nature of human pose configuration and to capture these characteristics across different scales and resolutions. The network is fractal in that it reflects the co-occurrance of inception and residual design at both the highest and lowest levels of abstractions. 

\subsection{Transfer Learning and Guided Training}

Nevertheless, training such deep networks has proven to be challenging \cite{erhan2009difficulty}. Significant efforts has been devoted to alleviate this problem. 
For instance, there has been another line of work in which a student network is trained from scratch to mimic the behavior of a much larger teacher network. Staring from Bucila et al.’s work \cite{buciluǎ2006model} and Hinton et al.’s more general Knowledge Distillation (KD) \cite{hinton2015distilling} approach, the knowledge transfer in learning process has gained a lot of research interest. 
In this paper, we consider a unique setting of the problem. Instead of transferring knowledge from teacher networks into a student network, we propose an external knowledge representation and projection framework to guide the training process of our deep neural network for human pose estimation. 
Specifically, we inject hand-designed features that are inferred from ground truth as external knowledge to aid the training of a highly complex network with deep structure and multiple loss functions. 
Inspired by \cite{ren2014face}, which proposed a locality principle to learn task-specific feature mapping for shape regression, we project the external knowledge with a learned feature mapping. 
The procedure involves domain adaptation and model training simultaneously. Since external knowledge is inferred from ground truth, it is inherently more reliable and effective than the outputs from a teacher network.

\section{Proposed Method}
\label{sec:proposed-method}

\begin{figure}[t]
	\begin{center}
		\includegraphics[width=1.0\linewidth]{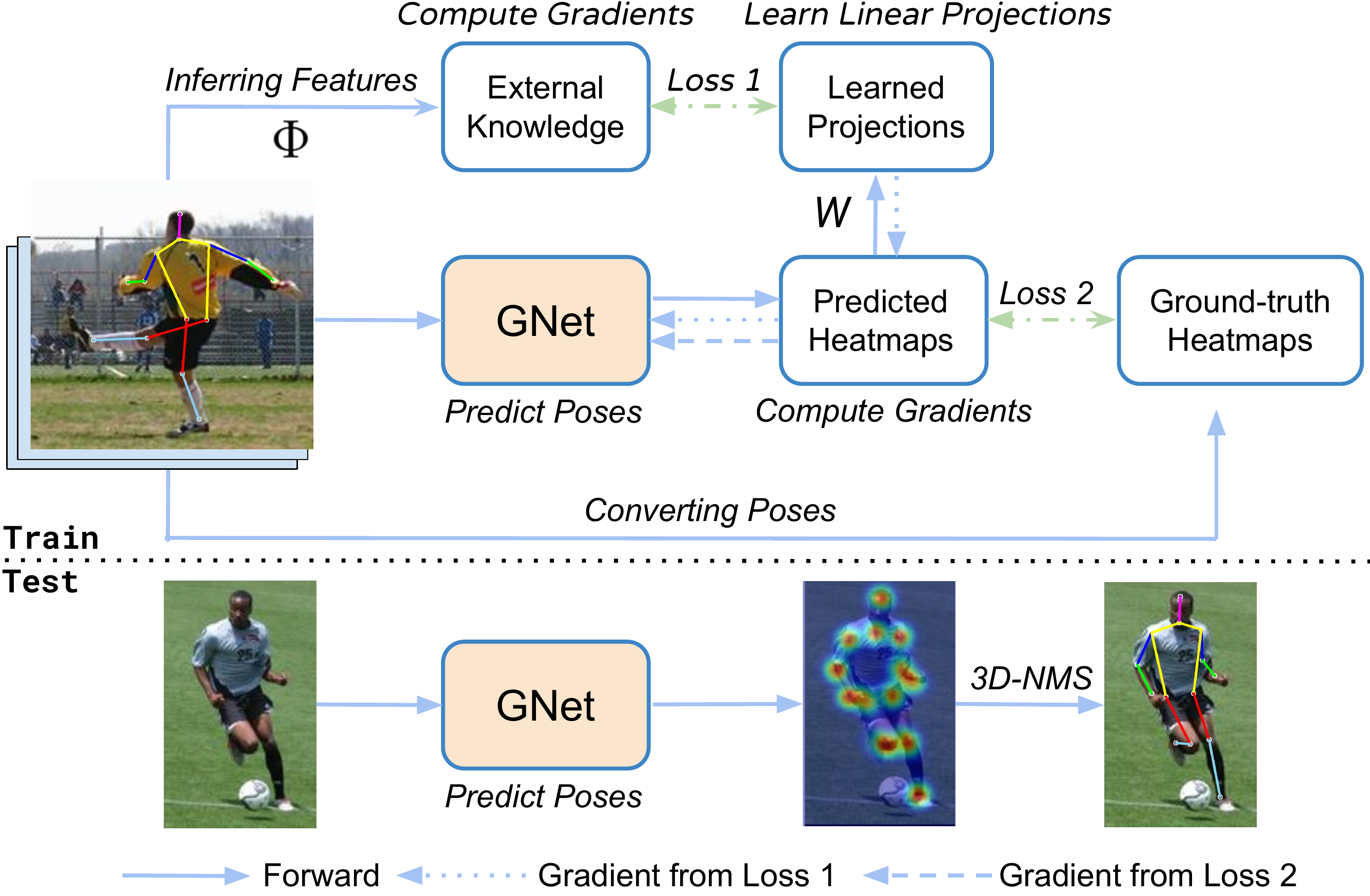}
	\end{center}
	\caption{\textbf{Framework of our proposed Guided Network (GNet)}. The projected knowledge affects the gradients propagated back to convolutional layers but they are not part of the network during deployment. By enforcing constraints with external knowledge injection, high-level information of long-range dependencies between image and multi-part cues that is hard to capture with implicit learning can be better learned under the guidance of mid-level knowledge projection.}
	\label{fig:framework}
\end{figure}

\subsection{Network Structure and Design}
\label{sec:fractal-network}

Human pose estimation methods using hand-crafted features or graphical structure models based on human knowledge lack the flexibility in learning and the potential to achieve great representation power. On the other hand, pure data-driven neural networks may not be able to capture sophisticated knowledge involved in human pose estimation.
In this work, we propose to represent and inject external human knowledge to guide the learning of deep neural networks (DNN), as illustrated in Fig. \ref{fig:intro}. 
Our major idea is that, by enforcing constraints and guidance with external knowledge injection, high-level information of long-range dependencies between image and multi-part cues, that are hard to capture with implicit learning, can be better learned under the guidance of mid-level knowledge projection.
As shown in Fig. \ref{fig:framework}, the projected knowledge affects the gradients propagated back to convolutional layers during training, but they are not part of the network during test.

We borrow the ideas from inception-residual networks \cite{szegedy2016inception} and propose to construct a basic inception-resnet module in replacement of convolutional layers for more robust feature representation. 
Hourglass network is first introduced in \cite{newell2016stacked} where features are processed across all scales by repetitve down-sampling and up-sampling and then consolidated to best capture various spatial relationships associated with the body. We introduce a modified version of the hourglass network with the proposed inception-resnet module. 
As shown in Figure \ref{fig:overview}, we use proposed inception-resnet modules and improved hourglass sub-networks to construct a fractal network to regress human pose images into heatmaps with no explicit graphical modeling. The network is fractal in that it has the same network configuration at all levels of analysis and abstractions.
This fractal network is designed to capture the multi-scale interdependence nature of human pose configuration and to represent these characteristics across different scales and resolutions.
In the inception network, we perform channel-wise concatenation of two tensors from different sources. This enforces the information represented by the features stored in these tensors to be complementary to each other. It encourages and directs these two sources to work on different concepts to produce a more robust union representation \cite{he2016deep}. In the 
Resnet model, we perform pixel-wise addition of two tensors with the same number of channels. From our experiments, we find that this network design allows us to train the network more effectively, since it enforces two separate tensors to be simultaneously accurate in order to render the expected outputs.

\begin{figure*}
	\begin{center}
		\includegraphics[width=0.9\linewidth]{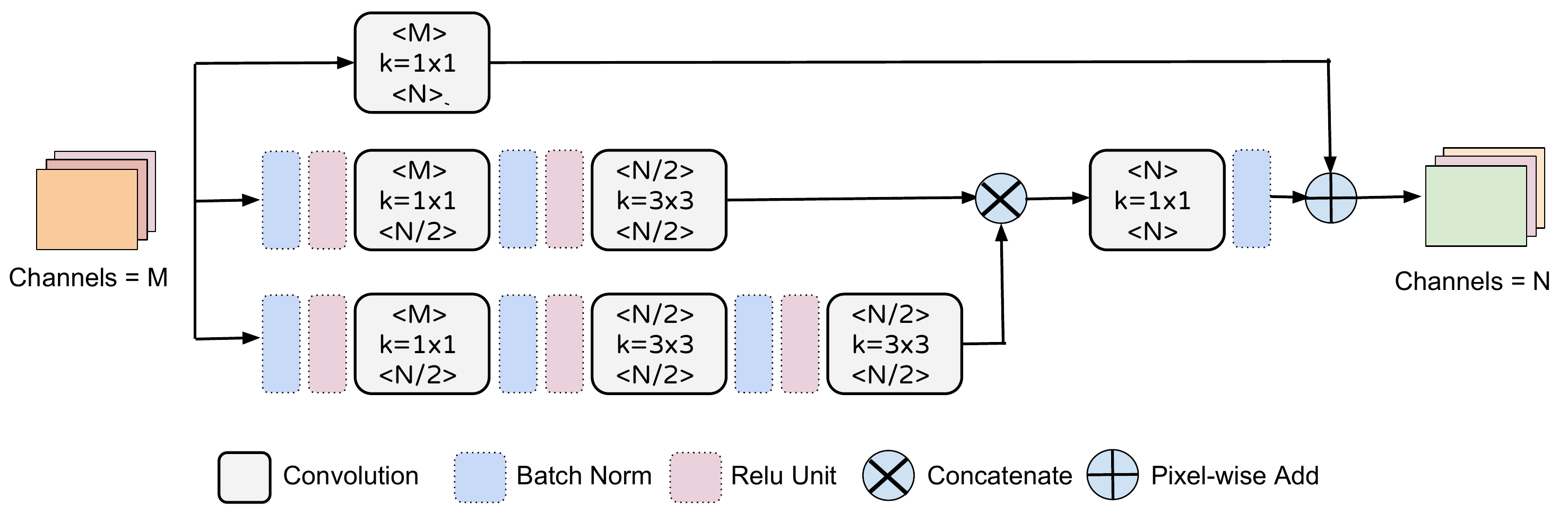}
	\end{center}
	\caption{\textbf{Basic module: Inception-resnet}. Convolution layers are padded such that the resolution of output is the same as that of the input. The benefit of this module is that the input and output are of uniform resolution while the depth of channels can be changed. The function of this basic module is to interpret the input information from one form to another, extracting features for another abstraction level with little loss of information quantity.}
	\label{fig:inception-resnet}
\end{figure*}

\subsection{Fractal Network with Inception-resnet Modules}
\label{sec:inception-resnet}

Our motivation in the fractal network design is that, we need the network to focus on various scales across human parts, and at each scale, the network should also have an overall understanding of this receptive field.
At higher levels, the network captures dependencies among various human parts. At lower levels, we use same fractal design to capture regional dependencies. It is essential to capture local dependencies in addition to local appearances. Because at a certain high-level scale, the receptive field may involve a human part as well as noises from other parts. These adjacent parts may be from the same or other persons. Therefore, local dependencies are helpful in providing more reliable features to higher-level networks.

\begin{figure}
	\begin{center}
		\includegraphics[width=1\linewidth]{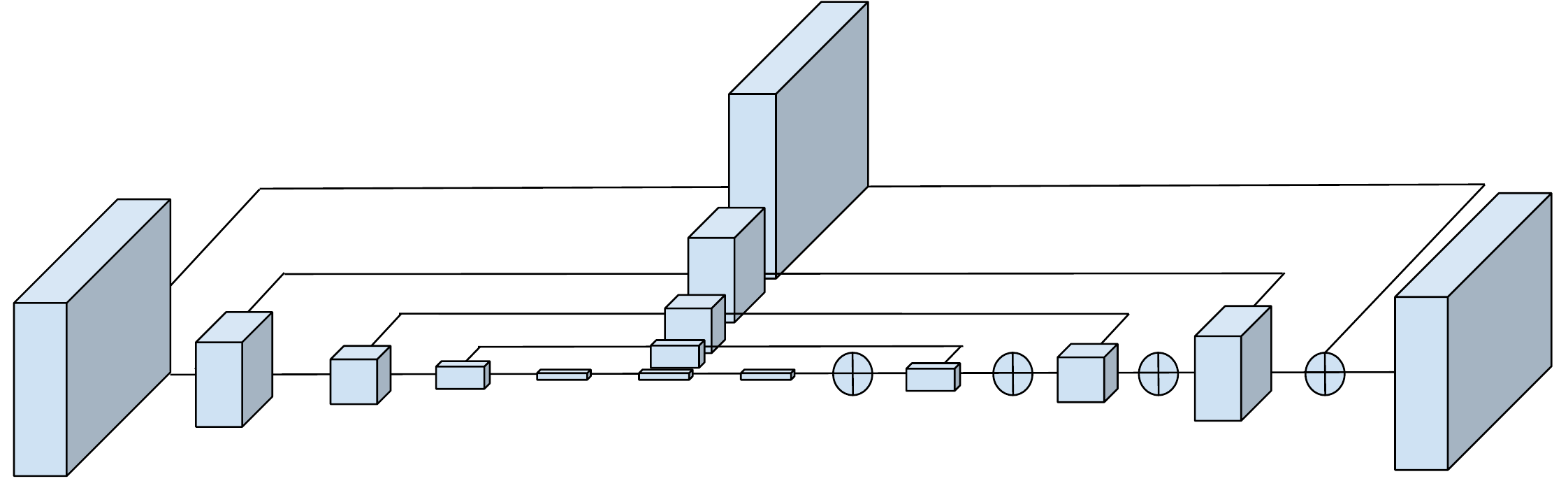}
	\end{center}
	\caption{An illustration of hourglass design proposed in \cite{newell2016stacked}. Pixel-wise addition fuses the information from two branches while keeping the input and output resolution uniform. The illustration gives an example of a 4-level hourglass.}
	\label{fig:hourglass-org}
\end{figure}

\begin{figure*}
	\begin{center}
		\includegraphics[width=0.9\linewidth]{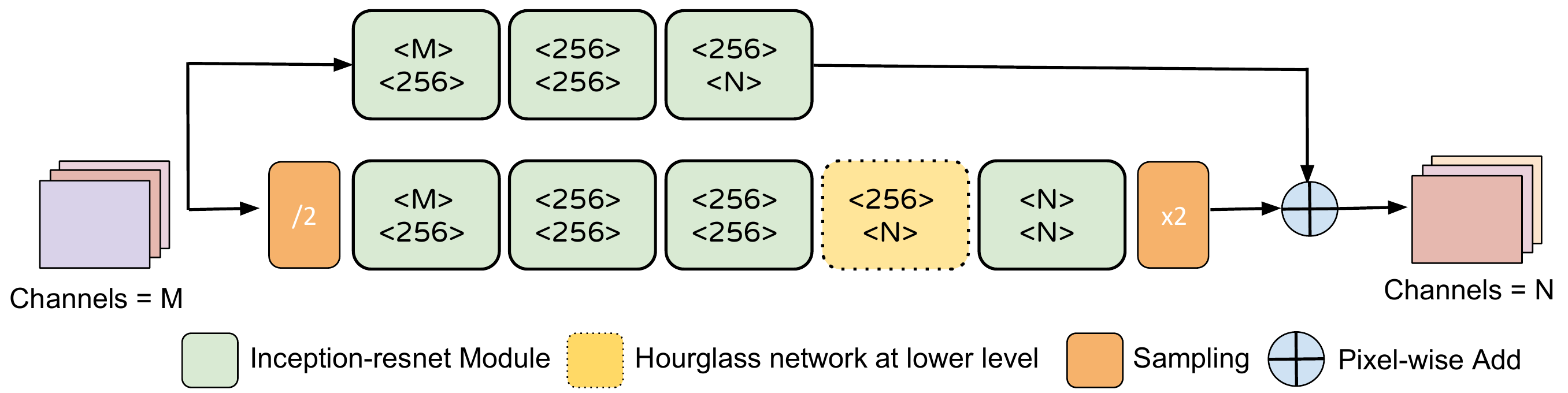}
	\end{center}
	\caption{\textbf{Improved hourglass sub-network}. While using inception-resnet as proposed common building block, we borrow the idea of hourglass design by down-sampling and then up-sampling the dataflow in one branch, maintaining the resolution of the other branch. The lowest level of the recursive hourglass in our network is an inception-resnet module.}
	\label{fig:hourglass}
\end{figure*}

The construction of inception-resnet module is shown in Figure \ref{fig:inception-resnet}. Based on the hourglass design proposed in \cite{newell2016stacked} shown in Figure \ref{fig:hourglass-org}, an improved version of hourglass network is developed in this work as a mid-level sub-network which also uses inception-resnet modules as the basic units, as illustrated in Figure \ref{fig:hourglass}.  
To combine the advantages of both inception and resnet design, we introduce the inception-resnet module as the basic building block to analyze local fields, while using an improved hourglass network to capture the global information of different parts. 

At the bottom level, we propose to use inception-resnet module as the basic structure unit of the network. It consists of convolutional layers, batch norm layers and relu units, with channel-wise concatenation and pixel-wise additions.
Convolution layers are padded such that the resolution of output is the same as that of the input. 
Although the concatenation of two branches maintains different level of information, the concatenated features across different channels need to be transformed and normalized by the subsequent convolutional layers. In the proposed inception-resnet module, the concatenation layer is followed by another convolutional layer with $1 \times 1$ kernels. 
The benefit of this module is that the input and output have the same resolution while the depth of channels can be flexible. 

At the sub-network level, we implement the recursive hourglass for $4$ levels as shown in Figure \ref{fig:hourglass-org}. In other words, it will process the image at four scales. The hourglass network is nested in itself. The first level of hourglass in our network is an inception-resnet module.
As illustrated in Figure. \ref{fig:hourglass}, we also borrow the idea of hourglass design by down-sampling and then up-sampling the data while using inception-resnet as proposed common building block. 
Pixel-wise addition fuses the information from two branches while keeping the input and output resolution the same.

At the top fractal network level, images of size $256 \times 256$ are down-sampled into the resolution of $64 \times 64$. Subsequently, inputs and outputs of all modules are of size $64 \times 64$, including the output heatmaps. 
The network captures and consolidates information across all scales of the image. 

\subsection{External Knowledge Representation}
\label{sec:guided-learning}

The fractal network is used to boost the data representation power of the deep neural network for human pose estimation. As the network grows deeper and more complicated, it requires careful attention to the training process. Furthermore, we recognize that the deep neural network is inherently an algebraic computing system, which might not be the most efficient way to capture the highly sophisticated human knowledge during pose estimation, for example those highly coupled geometric constraints and interdependence among body joints. To address these two issues, in this work, we propose to encode and inject external knowledge into the fractal network to guide the training process of the network using learned projections, enforcing a prior during the training process.

In this work, we propose to inject the geometric representation of knowledge into the heatmap layer of the network. 
Since the heatmaps to be predicted are correlated to each other as they largely share parameters on former layers, the constraint on one heatmap influences the parameters of these layers and therefore having an impact on the training of other heatmaps. We observe that intermediate layers in our network are low and mid-level visual features; higher-level semantic features are hard to locate and explicitly interpret. 
The predicted heatmaps are easier to enforce the external knowledge and constraints upon. 
During the training process, the external knowledge and its visual representations are projected into the background and keypoint heatmaps using a projection matrix. 
We find that this type of knowledge-guided learning inherently enforces long-range dependencies and configurations among human joints, while leaving the flexibility of representation to the depth of the network, the quality and quantity of training data. In the following, we explain the proposed method in more detail.

During the training process, the external knowledge representation module illustrated in Figure. \ref{fig:intro} has access to the original training sample image and its ground-truth joint locations.

Specifically, during feature mapping which is denoted as $\Phi$, we perform \textit{Hough Transform} on each line traversing two separate joints denoted as $(u_{i}, v_{i})$ and $(u_{j}, v_{j})$. In Hough space, each line is represented by a coordinate $(\theta, \rho)$.

\begin{equation}
\begin{cases} 
\theta  =  \arctan (\dfrac{x_{i} - x_{j}}{y_{j} - y_{i}}) \\ 
\rho =  x_{j} \times \cos(\theta) + y_{j}   \times \sin(\theta) 
\end{cases} 
\end{equation}

\begin{figure}[t]
	\begin{center}
		\includegraphics[width=1.0\linewidth]{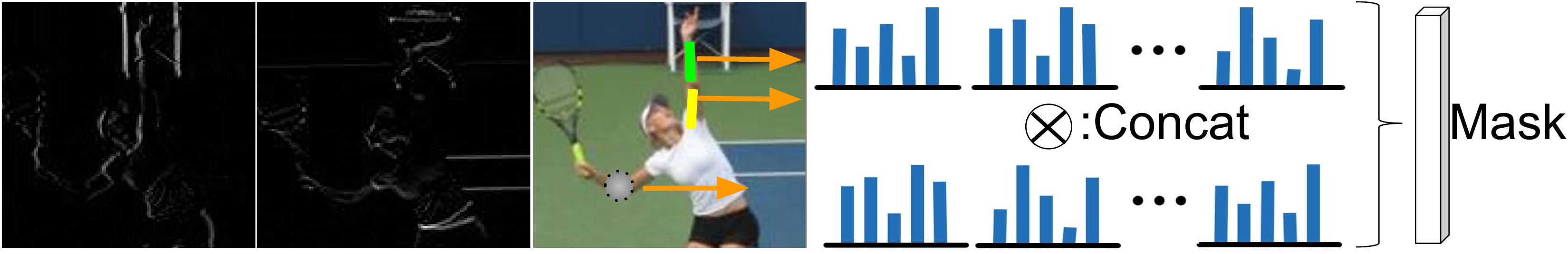}
	\end{center}
	\caption{We encode image descripors around each pair of adjacent joints in order to capture visual features to compensate spatial dependencies. The features are concatenated and normalized for the external knowledge representation.}
	\label{fig:feat}
\end{figure}

In order to represent the information in a less crisp manner, we convert the coordinates into a normalized vector representation. To incorporate the inherent learning of geometrical features such as angles and distance, we also inject the joint locations alongside each line. Based on the visibility of each joint, the line traversing it is encoded with the number of visible joints.

In addition to encoding geometric features, we encode image descriptors such as \textit{Histogram of Gradients (HOG)} \cite{dalal2005histograms} around each pair of adjacent joints in order to capture visual features to compensate spatial dependencies. 
While we preserve the flexibility of deep convolutional features that automatically learn visual semantics, we use hand-crafted features as guidance of the learning by enforcing a strong prior during the training of the neural network. We noticed that human joints may connect to those from an adherent person, even though the ground truth joints are not self-occluded or object-occluded. We believe HOG features are helpful in observing edges and therefore distinguishing real and false limbs.
The injected features for pairs of limbs impose a strong prior during the training, preventing human part keypoint from connecting to noises, e.g., keypoint from other people in the background that is not cropped out for the target person, which is helpful in the learning of body part interdependencies. 
As illustrated in Figure \ref{fig:feat}, the features are concatenated and normalized for the external knowledge representation. For self-occluded and object-occluded joints, we mask corresponding features with zeros.
Specifically, we follow the traditional HOG feature extraction schemes, applying filters $D_{x} = [-1 \quad 0 \quad 1]$ and $D_{y} = [1 \quad 0 \quad -1]^{T}$ horizontally and vertically to generate gradient maps $I_{x}$ and $I_{y}$. Instead of scanning a window for blocks and cells over the image which is done in traditional ways, we locate limbs based on meta-data from the training set and extract histogram of gradients for such regions. 
The magnitude and orientation of the gradient are respectively computed by:
\begin{equation}
|G| = \sqrt{I_{x}^{2} + I_{y}^{2}} 
\end{equation}
and
\begin{equation}
\varphi = \arctan\dfrac{I_{y}}{I_{x}}
\end{equation}
We use $8$ bins for the pooling, followed by block normalization (L2-norm) to mitigate the effect of unbalanced area of regions:
\begin{equation}
f = \dfrac{v}{\sqrt{||v||_{2}^{2}} + e^{2}}
\end{equation}
Where $e$ is a very small number.

\subsection{Knowledge Projection into the Deep Neural Network}   

In favor of decoding the abstract external knowledge in higher-dimensional space, we afford 2 fully-connected (FC) layer and 3 convolutional layers for the for geometric features and edge features, between the projection representation and the injected knowledge to learn linear projection $W$, which will be removed during testing as it is undesirable to keep redundant layers.

We inject external features as knowledge $K$ via global feature mapping function $\Phi$ and learn a global linear projection $W$ by minimizing the loss from the knowledge projection layer:
\begin{equation}
\mathcal{L}_{KP} = ||K - W \times H_{J}||_{2}^{2} + \beta \times ||W||_{2}^{2}
\end{equation}
where the first term is the regression target, the second term is a $L2$ regularization on $W$, and $\beta$ controls the regularization strength. Regularization is necessary because the dimensionality of the features is very high. Since the objective function is quadratic with respect to $W$, we can always reach its global optimum \cite{ren2014face}.

Specifically,  
we enforce two 
loss functions, one for injected geometric features and one for limb-wise edge features.  (1) The ground truth heatmap is convolved by 1x1 kernels, outputting 8 channels of maps. It is padded such that the resolution does not change. A fully connected layer with an output of 224-dimensional geometric feature is added to the convolutional layer. We add L2 loss (weighted by 0.05) for the geometric features and the inferred features from the ground truth. (2) We branch out the 3rd inception-residual module at the early stage and feed its output to a series of convolutional layers with 1x1 kernels. The numbers of output channels are scaled twice by a factor of 1/2 until it reaches 32 channels, followed by a fully connected layer. We add L2 loss (weighted by 0.05) for injected edge features and the inferred edge features.

We denote the pixel location of the $j$-th anatomical landmark (which we refer to as human joint), $Y_j \in \mathcal{Z} \subset \mathbb{R}^{2}$, where $\mathcal{P}$ is the set of all $(u, v)$ pixel locations in image coordinate system. Our goal is to predict the image locations $Y = (Y_{1}, ... Y_{J})$ for all $J$ joints.
The output heatmaps are of size $J \times 64 \times 64$, denoted by $H = (H_{1}, ..., H_{J})$, which are predicted beliefs for assigning a pixel location to each joint $Y_j = p, \forall z \in \mathcal{P}$, producing belief scores $S_{j}$ for all pixels in the heatmap of joint $j$:  

\begin{equation}
H_{j}(p) \leftarrow S_{j}(Y_{j} = p)
\end{equation}

\begin{figure*}[t]
	\begin{center}
		\includegraphics[width=1.0\linewidth]{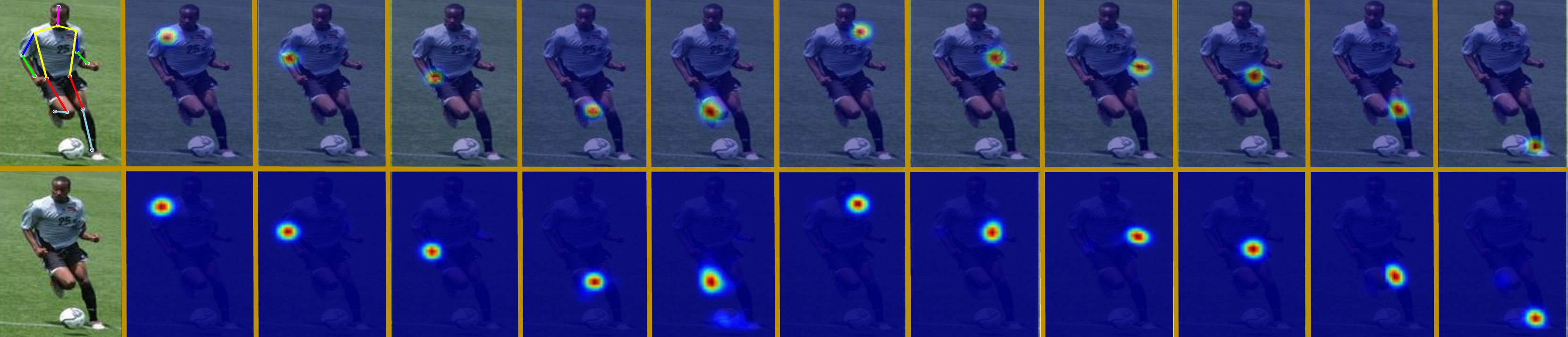}
	\end{center}
	\caption{Example output produced by our network. On the top-left we see the final pose estimate provided by NMS across all heatmaps. Elsewhere we show sample heatmaps: (1) The first row shows the final part regression heatmap results; (2) the second row shows the preliminary part regression results from the intermediate supervision layer. The heatmaps from the first row have finer predictions than the second row, especially the heatmap for the right foot, where the preliminary prediction renders belief scores for the soccer ball as well. }
	\label{fig:heatmaps}
\end{figure*}

In our experiments, we regress RGB-channel images into a set of $15$ heatmaps, $14$ of which are human joints while the other one as the background. The heatmaps are then suppressed into joint locations $Y$ with our proposed 3D-NMS algorithm specially designed for human pose estimation.
During training, we provide ground truth heatmaps for each joint by creating Gaussian peaks at ground truth locations. The cost function $\mathcal{L}_{f}$ we aim to minimize for the fractal network is given by:

\begin{equation}
\mathcal{L}_{f} = \sum_{j \in J}||H_{j}(p) - H_{j}^{*}(p)||_{2}^{2}
\end{equation}

The overall loss for training is a weighted combination of heatmap cost and projection matrix fitness provided by knowledge-guided learning, with a control parameter on how much guidance should be imposed. 
The overall network is then trained to minimize the following joint loss function:
\begin{equation}
W_{f}^{*} \leftarrow \argmin_{W_{f}}(\lambda \times \mathcal{L}_{KP} + (1 - \lambda) \times \mathcal{L}_{f})
\end{equation}
where $\mathcal{L}_{KP}$ and $\mathcal{L}_{f}$ are loss from knowledge projection layer and the fractal network loss, $\lambda$ is the weight parameter decaying during training, 
and $W_{f}^{*}$ is the trained parameters in the fractal network.

The output of knowledge projection layer will guide the training of fractal network by generating a strong and explicit gradient applied to backward path to the injection layer in the following form:
\begin{equation}
\Delta W_{f, i} = - \lambda \cdot \dfrac{\partial \mathcal{L}_{KP}}{\partial W_{f,i}}
\end{equation}
Where $W_{f, i}$ is the weight matrix of injection layer in fractal network.
Note that the network update only occurs during training. During testing, the knowledge representation and projection modules are removed.

\subsection{Cross-Heatmap Non-Maximum Suppression}

In this work, we introduce a novel pose non-maximum suppression (NMS) algorithm specially designed for human pose estimation.
Our experiments in Section \ref{sec:component-analysis} show that employing pose-NMS consistently render better predictions for all models across iterations on both MPII \cite{andriluka20142d} and LSP \cite{johnson2010clustered} datasets.
Instead of finding the maximum value at pixel-level to predict joint location as in \cite{newell2016stacked, belagiannis2016recurrent, wei2016convolutional, bulat2016human}, we detect blobs with high responses in each heatmap. Basically, we gather blobs from all heatmaps for suppression. We first find the blob with maximum response, then suppress other blobs from the same heatmap, and blobs from other heatmaps very close to this blob in image coordinate system. We repeat this procedure until all blobs are removed. The suppression takes place in image coordinate system and channel-wise $(u, v, c)$, therefore called cross-heatmap NMS.

\section{Summary of Training and Testing Procedures}

We summarize our training and testing procedures in Algorithm \ref{algorithm:training} and \ref{algorithm:testing}, respectively. 
There exist around 250 convolutional layers in the original hourglass network, while the proposed network with inception-resnet modules consist of over 300 convolutional layers. The network for training the proposed network has an additional cost with 1 external feature extraction module, 2 fully connected layers, 3 convolutional layers and 2 additonal loss layers. In our implementation, it takes the hourglass network  an average of 47ms to feed forward with a single Pascal TITAN X GPU. In comparison, the feed forward time of the proposed network with inception-resnet modules during testing is 62ms.

\begin{algorithm}[h]
	\SetAlgoLined
	
    \SetKwInOut{Input}{input}
    \SetKwInOut{Output}{output}
    
    \Input{A set of RGB images $I$ and corresponding ground truth joint coordinates $J$}
    \Output{Trained weights $W_{f}^{*}$ for the Fractal Network, $W_{KP}^{*}$ for the knowledge projection layers}

	Initialize DNN with fractal network and knowledge projection layers\;
	\For {$k$ epoches}
	{
		\For {mini-batch in $I$}
		{
			Compute external knowledge representation: $K \longleftarrow \{ I_{n}, J_{n} \} $ \;
		Back-propagate w.r.t $W_{f}, W_{KP}$ \;
		$ \{ W_{f}^{\prime}, W_{KP}^{\prime} \} \leftarrow \argmin_{W_{f}}(\lambda \times \mathcal{L}_{KP} + (1 - \lambda) \times \mathcal{L}_{f} )$ \;
	    }
	}
    return $\{ W_{f}^{*}, W_{KP}^{*} \}$

\caption{Summary of Procedures: Training Phase}
\label{algorithm:training}
\end{algorithm}

\begin{algorithm}[h]
	\SetAlgoLined
	
	\SetKwInOut{Input}{input}
	\SetKwInOut{Output}{output}
	
	\Input{A set of RGB images $I$ and a fractal network with trained weights $W_{f}^{*}$}
	\Output{A set of predicted joint coordinates $J$ in the same image coordinate system}
	
	initialize network only with fractal network layers $W_{f}^{*}$ \;
	\While{not at end of this image set}{
		Load image $I_{i}$ \;
	    Forward the network: $J_{i} \leftarrow \{ W_{f}^{*}, I_{i} \} $ \;
	    
    }
    return $J$
\caption{Summary of Procedures: Testing Phase}
\label{algorithm:testing}
\end{algorithm}

\section{Experimental Results}

For comprehensive experimental analysis, we will first introduce the datasets, evaluation criteria and implementation details. Then we will present quantitative evaluations on benchmark datasets. Finally, diagnostic experiments, algorithm performance analysis and dicussions are provided for further analysis.

\label{sec:experiments}
\subsection{Datasets and Criteria}
\subsubsection{Datasets}
We evaluate the proposed method on two widely used benchmarks: MPII Human Pose \cite{andriluka20142d} and extended Leeds Sports Poses (LSP) \cite{johnson2010clustered}. The MPII Human Pose dataset includes about $25$K images with $40$k annotated poses. The images are collected from YouTube videos covering daily human activities with highly articulated human poses. The LSP dataset with extended training data consists of 11K training images and 1K testing images from sports activities.

\subsubsection{Criteria}
There are three criteria used in the experiments to evaluate the performance of the proposed human pose estimation approach: Percentage of Corrected Parts (PCP) \cite{yang2013articulated, ferrari2008progressive, eichner20122d}, Percentage of Detected Joints (PDJ) \cite{toshev2014deeppose, sapp2013modec, yang2013articulated}, and Percentage of Corrected Keypoints (PCK) \cite{yang2013articulated}.

\paragraph{PCP}
A widely-used criterion for human pose estimation is PCP which evaluates the localization accuracy of body parts (sticks of skeleton). It requires the estimated part end points must be within half of the part length from the ground truth part end points. As pointed by Yang and Ramanan \cite{yang2013articulated}, some previous work requires only the average of theendpoints of a part to be correct (PCP-average), rather than both endpoints (PCP-strict). Moreover, the early PCP implementation \cite{ferrari2008progressive} selects the best matched output without penalizing false positives. In all our experiments, we adopt the strictest measure, i.e., PCP-strict with single output, if not specially specified. For more detailed descriptions on PCP, it is recommented to refer to \cite{ferrari2008progressive} and \cite{yang2013articulated}.
\paragraph{AUC}
Though PCP is the initially preferred criterion for evaluation, it has the drawback of penalizing shorter limbs, such as lower arms. Thus PDJ is introduced \cite{toshev2014deeppose, sapp2013modec} to measure the detection rate of body joints, where a joint is considered to be detected if the distance between the detected joint and the true joint is less than a fraction of the torso diameter. The torso diameter is usually defined as the distance between opposing joints on the human torso, such as left shoulder and right hip \cite{toshev2014deeppose}. The Area Under Curve (AUC) can be used as the overall evaluation of the PDJ curve. In the following experiments, we report AUC as our PDJ performance.
\paragraph{PCK}
The PCK measure is very similar to the PDJ criterion. The only difference is that the torso diameter is replaced with the maximum side length of the external rectangle of ground truth body joints. For full body images with extreme pose (especially when the torso becomes very small), the PCK may be more suitable to evaluate the accuracy of body part localization.

In our experiments, we follow the official benchmark evaluation protocals 
\footnote{http://human-pose.mpi-inf.mpg.de/\#evaluation}. 
Official benchmark on MPII dataset adopts PCKh (using portion of head length as reference) at $0.5$, while official benchmark on LSP dataset adopts both PCP and PCK at $0.2$. LSP benchmark provide comparisons on both Observer-Centric (OC) and Person-Centric (PC) evaluations, of which the most widely adopted evaluation protocal is PCK-PC. In addition, both benchmarks adopt AUC scores.

\subsection{Implementation Details}

\begin{table*}[t]
	\small
	\tabcolsep=0.22cm
	\begin{center}
		\begin{tabular}{|l|c c c c c c c c|}
			\hline
			Method &Head &Sho. & Elb. & Wri. & Hip & Knee & Ank.  &Total \\
			\hline\hline
			Hourglass &97.0  & 93.0  & 88.8  & 85.6  & 92.2  & 93.0 & 90.9 & \textbf{91.5}  \\
			Ours (no guidance)  & 97.9  & 93.2  & 89.1  & 86.4  & 94.5  & 93.8 & 92.9 & \textbf{92.6} \\
			Ours (with guidance)  & 98.2  & 94.4  & 91.8  & 89.3  & 94.7  & 95.0 & 93.5 & \textbf{93.9} \\
			
			\hline
			Plain testing   
			& 97.4  & 92.7  & 88.8  & 86.7  & 92.2  & 93.8 & 92.2 & 92.0 \\
			\quad\quad + flipping  & 97.7  & 93.3  & 90.4  & 87.5  & 93.2  & 94.2 & 92.8 & 92.7 \\
			\quad\quad\quad + scaling  & 98.1  & 93.7  & 91.3  & 88.7  & 94.0  & 94.6 & 93.2 & 93.4  \\
			\quad\quad\quad\quad + 3D-NMS   & 98.2  & 94.4  & 91.8  & 89.3  & 94.7  & 95.0 & 93.5 & \textbf{93.9} \\
			\hline
		\end{tabular}
	\end{center}
	
	\caption{\textbf{Component analysis} on the LSP Dataset of PCK@0.2 score. Note that numbers in bold indicate the method has employed all techniques during testing.}
	\label{table:component-analysis}
\end{table*}

\subsubsection{Data Augmentation}
We crop the images with the target human centered at the images with roughly the same scale, and warp the image patch to the size $256 \times 256$. Then, we randomly rotate ($\pm 30^{\circ}$) and flip the images, perform random re-scaling ($0.75$ to $1.25$) and color jittering to make the model more robust to scale and illumination changes. 

\subsubsection{Experimental Settings}
We use a modified version of \textit{Caffe}\cite{jia2014caffe} that produces three kinds of outputs from the data layer: the augmented image, the corresponding transformed ground truth heatmaps, and the injected knowledge for the augmented image. The knowledge projection is switched off during testing.  
We train our model using the initial learning rate of $2.5 \times 10^{-4}$ . The parameters are optimized by RMSprop \cite{tieleman2012lecture} algorithm. We divide the learning rate by 2 when the validation set hits plateaus. The minimum learning rate is set to $10^{-6}$.
We use 4 Pascal TITAN GPUs to train the model on the merged dataset of MPII and extended LSP for over $300$ epochs, and adopt Tompson's validation split for the MPII dataset used in \cite{tompson2014joint} to monitor the training process.
The same model is used for the testing of both MPII and LSP test sets.
According to \cite{pinheiro2014recurrent}, there is a prior towards the background that forces the network to converge to zero. It is therefore important to weight the gradient responses so that there is an equal contribution to the parameter update between the foreground and background heatmap pixels. In our training process, we weight the foreground and background by $20:1$. 
The neural network takes the cropped images patches or ROI of the images as inputs. However, there exists such situation where the cropped patches or ROI contains limbs from other persons. In this case, our ground truth simply ignores other limbs. For example, any region that is not from the keypoints of the target person is considered as background heatmap in the ground truth. Since the target person is always centered in the cropped image or ROI, it enforces a prior during training. Therefore, limbs from other persons are usually of lower response, reflected by the predicted heatmaps. 

\subsubsection{Inference}
During testing, we follow the standard routine to crop image patches with the given rough position and the scale of the test human for MPII dataset. For the LSP dataset, we use image size as the rough scale, and image center as the rough position of the target human to crop the image patches.
Before feeding into the neural network, we further pre-process images with normalization and pixel-wise subtraction by estimated mean value. 
All the experimental results are produced from the original and flipped image pyramids with 2 scales (1 and 0.75). Note that we swap heatmaps of left and right limbs before merging corresponding heatmaps for each joint. The merged heatmaps are transformed into joint coordinates by the proposed cross-heatmap non-maximum suppression method. 
The feed forward time of the network during testing is 62ms with a single Pascal TITAN X GPU.

\subsection{Benchmark Evaluation}

We use the Percentage Correct Keypoints (PCK) \cite{yang2013articulated} metric for comparisons on the LSP dataset, and the PCKh measure \cite{andriluka20142d}, where the error tolerance is normalized with respect to head size, for comparisons on the MPII Human Pose dataset.
We train our model by adding the MPII training set to the extended LSP training set with person-centric (PC) annotations, which is a standard routine \cite{belagiannis2016recurrent, pishchulin2016deepcut, insafutdinov2016deepercut, wei2016convolutional}. 

\subsubsection{Results on the MPII Human Pose Dataset}

\paragraph{AUC}
The AUC score of our network for MPII dataset is $63.6$.

\paragraph{PCKh@0.5}

Table \ref{table:MPI-results} reports the comparison of the PCKh performance of our method and previous state-of-the-art at a normalized distance of $0.5$.
Our total PCKh-0.5 score achieves state of the art performance at 91.2\%. We apply all techniques described in Section. \ref{sec:component-analysis} during testing. Note that we test at same multiple scales (1 and 0.75) as that used on LSP dataset, which may not be ideal. While cropping the images with the given scale of MPII dataset, for some images the feet are cropped out, therefore suffering a comparatively lower detection rate for ankles.

\begin{table*}[t]
	\centering 
	\footnotesize
	\tabcolsep=0.2cm
	\begin{tabular}{l c c c c c c c c }
		\hline 
		Method & Head & Sho. & Elb. & Wri. & Hip & Knee & Ank. & Total \\ [0.1ex]
		\hline 
		Ours  &98.1  & 96.3  & 92.2  & 87.8  & 90.6  & 87.6 & 82.7 & \textbf{91.2} \\
		\hline
		Newell et al., ECCV'16\cite{newell2016stacked} &{98.2} &96.3 &91.2 &87.1 &90.1 &87.4 &83.6 &90.9 \\
		Bulat\&Tzimiropoulos, ECCV'16 \cite{bulat2016human} &97.9 &{95.1} &89.9 &85.3 &89.4 &85.7 &81.7 &89.7\\  
		Wei et al., CVPR'16 \cite{wei2016convolutional} &97.8 &95.0 &88.7 &{84.0} &88.4 &82.8 &79.4 &88.5 \\  
		Insafutdinov et al., ECCV'16 \cite{insafutdinov2016deepercut} &96.8 &95.2 &89.3 &84.4 &88.4 &83.4 &78.0 &88.5 \\  
		Rafi et al., BMVC'16 \cite{rafi2016efficient} &97.2 &93.9 &86.4 &81.3 &86.8 &80.6 &73.4 &86.3 \\  
		Gkioxary et al., ECCV'16 \cite{gkioxari2016chained} &96.2 &93.1 &86.7 &82.1 &85.2 &81.4 &74.1 &86.1 \\  
		Lifshitz et al., ECCV'16 \cite{lifshitz2016human} &97.8 &93.3 &85.7 &80.4 &85.3 &76.6 &70.2 &85.0 \\  
		Pishchulin et al., CVPR'16 \cite{pishchulin2016deepcut} &94.1 &90.2 &83.4 &77.3 &82.6 &75.7 &68.6 &82.4 \\
		Hu\&Ramanan, CVPR'16 \cite{hu2016bottom} &95.0 &91.6 &83.0 &76.6 &81.9 &74.5 &69.5 &82.4 \\  
		Tompson et al., CVPR'15 \cite{tompson2015efficient} &96.1 &91.9 &83.9 &77.8 &80.9 &72.3 &64.8 &82.0  \\  
		Carreira et al., CVPR'16 \cite{carreira2016human} &95.7 &91.7 &81.7 &72.4 &82.8 &73.2 &66.4 &81.3 \\  
		Tompson et al., NIPS'14 \cite{tompson2014joint} &95.8 &90.3 &80.5 &74.3 &77.6 &69.7 &62.8 &79.6 \\  
		Pishchulin et al., ICCV'13 \cite{pishchulin2013strong}  &74.3 &49.0 &40.8 &34.1 &36.5 &34.4 &35.2 &44.1 \\
		\hline	
		\hline
	\end{tabular}
	\caption{Comparisons of PCKh@0.5 score on the MPII test set.}
	\label{table:MPI-results}
\end{table*}

\subsubsection{Results on the Leeds Sports Pose Dataset}

\paragraph{AUC}
The AUC score of our network for LSP dataset is $69.1$.

\paragraph{PCK@0.2}

\begin{figure}[h]
	\begin{center}
		\includegraphics[width=1.0\linewidth]{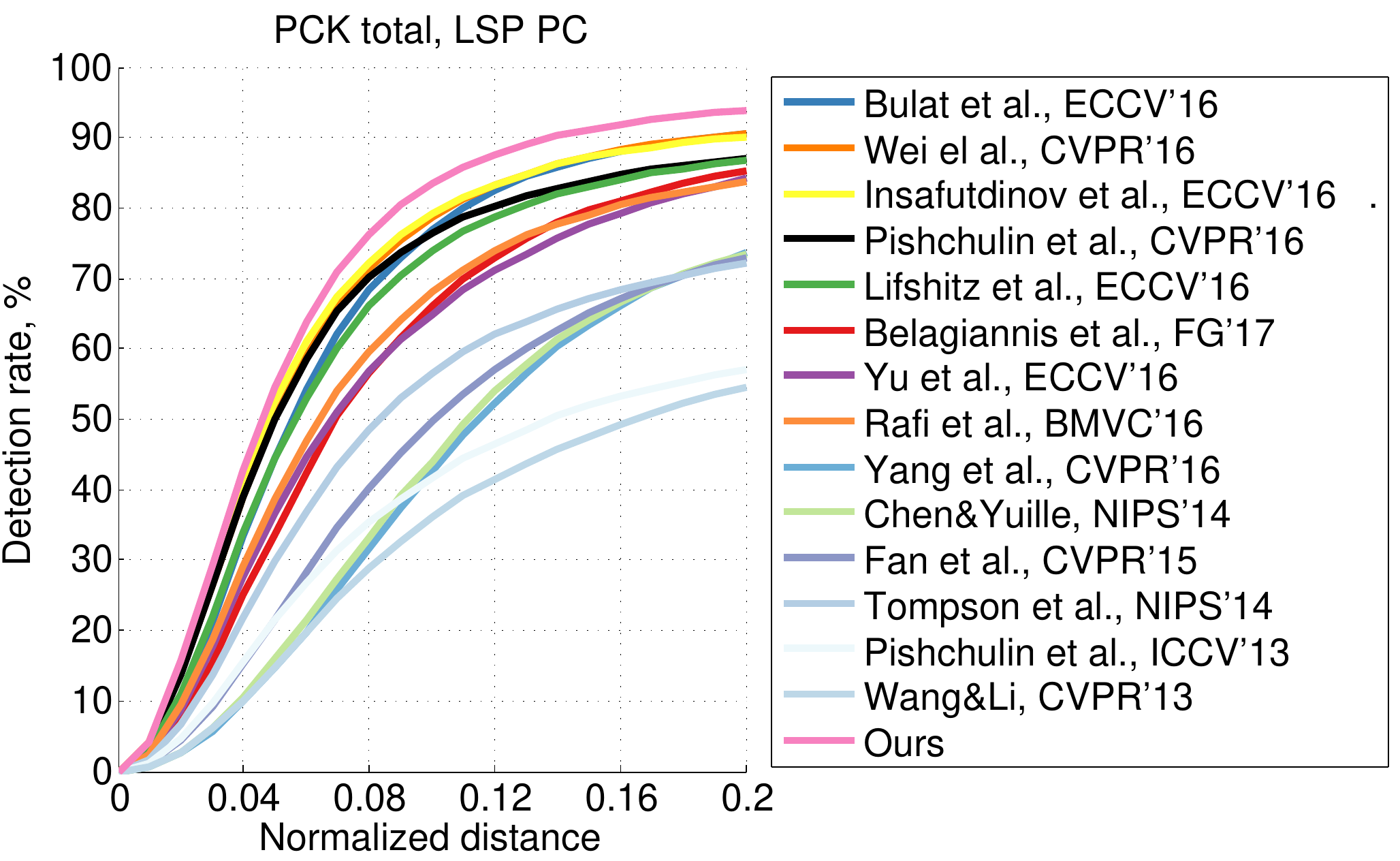}
	\end{center}
	\caption{Person-Centric (PC) PCK curves on the LSP test set. Ours is on top.}
	\label{fig:pck}
\end{figure}

Table \ref{table:LSP-results} reports the PCK at threshold of $0.2$, and Fig. \ref{fig:pck} exhibits PCK over various thresholds. Our approach achieves state-of-the-art performance with PCK value of 93.9\%, and outperforms all existing methods on each body part prediction. 

\begin{figure*}[t]
	\centering 
	\begin{center}
		\includegraphics[width=0.8\linewidth]{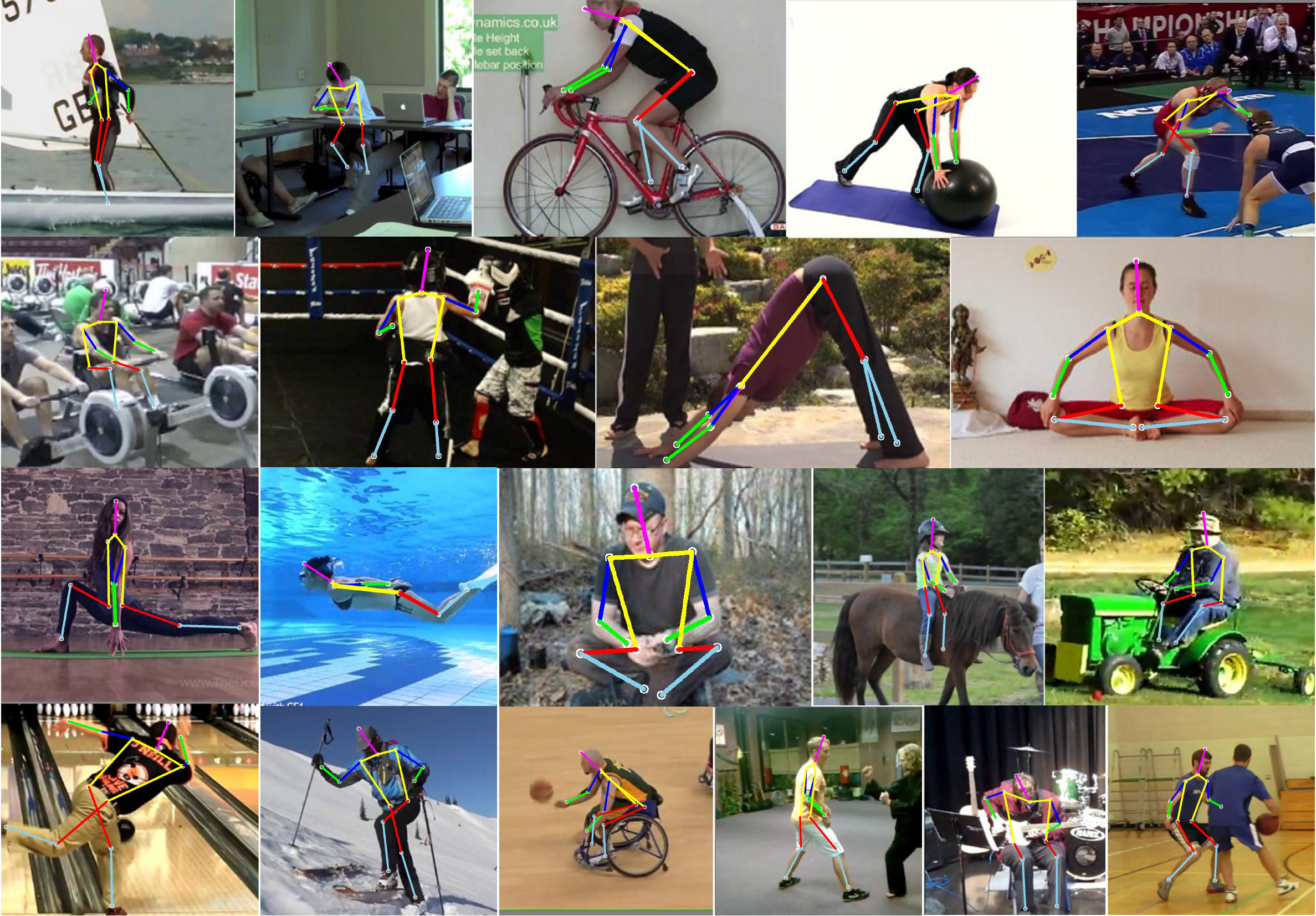}
		\caption{Qualitative results on the MPII test set.}
	\end{center}
	\label{fig:mpii}
\end{figure*}

\begin{figure*}[h]
	\begin{center}
		\includegraphics[width=0.8\linewidth]{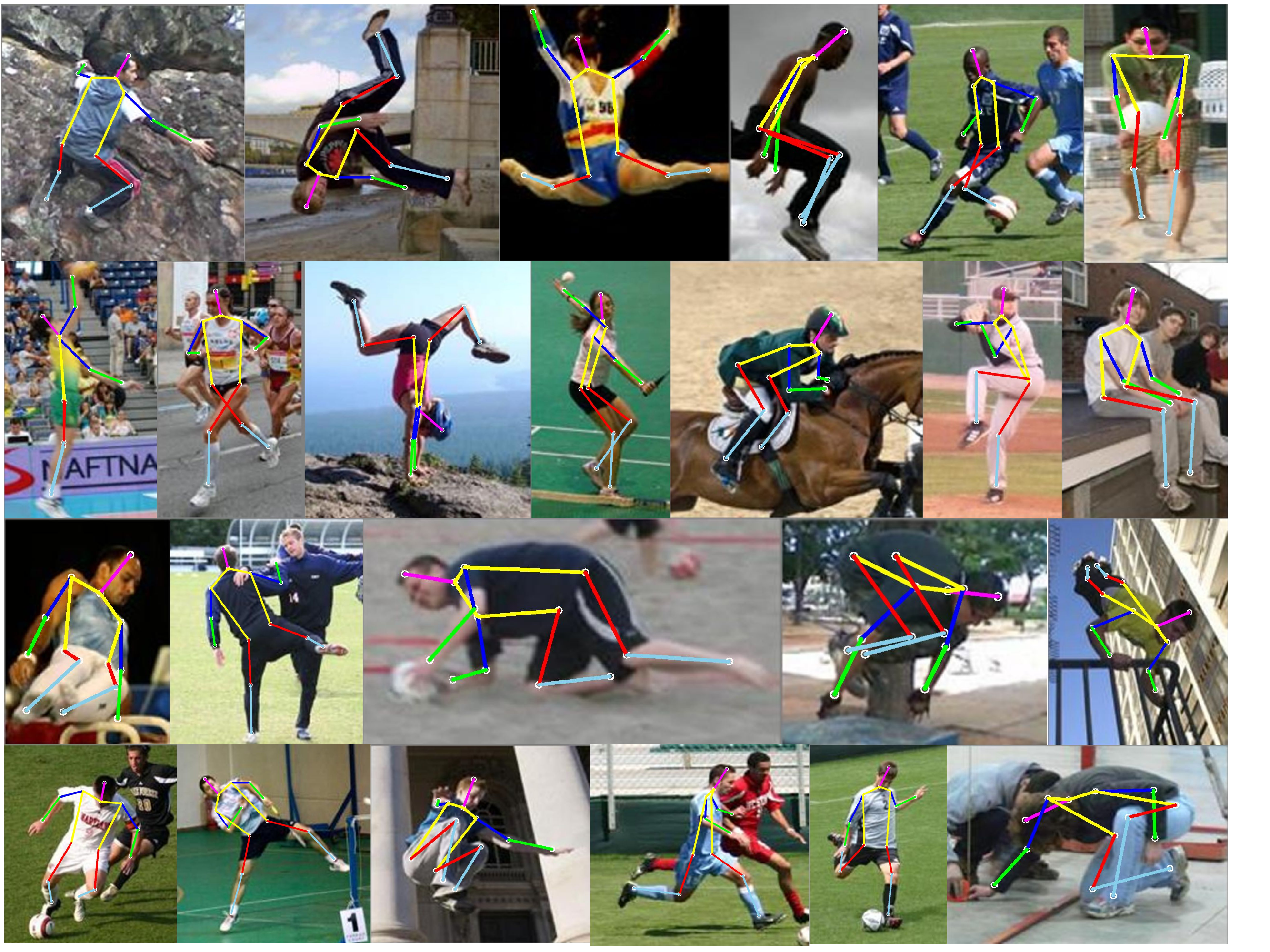}
	\end{center}
	\caption{Qualitative results on the LSP test set.}
	\label{fig:lsp}
\end{figure*}

\begin{table*}[t]
	\footnotesize
	\tabcolsep=0.2cm
	\begin{center}
		\begin{tabular}{l c c c c c c c c}
			\hline 
			Method & Head & Sho. & Elb. & Wri. & Hip & Knee & Ank. & Total \\ [0.2ex]
			\hline
		    Ours & 98.2  & 94.4  & 91.8  & 89.3  & 94.7  & 95.0 & 93.5 & \textbf{93.9} \\
			\hline
			
			Bulat\&Tzimiropoulos. ECCV'16 \cite{bulat2016human},  &97.2 &92.1 &88.1 &85.2 &92.2 &91.4 &88.7 &90.7 \\
			Wei et al. CVPR'16 \cite{wei2016convolutional}, &97.8 &92.5 &87.0 &83.9 &91.5 &90.8 &89.9 &90.5 \\
			Insafutdinov et al. ECCV'16 \cite{insafutdinov2016deepercut}, &97.4 &92.7 &87.5 &84.4 &91.5 &89.9 &87.2 &90.1 \\
			Pishchulin et al. CVPR'16 \cite{pishchulin2016deepcut},  &97.0 &91.0 &83.8 &78.1 &91.0 &86.7 &82.0 &87.1 \\
			Lifshitz et al. ECCV'16 \cite{lifshitz2016human}, &96.8 &89.0 &82.7 &79.1 &90.9 &86.0 &82.5 &86.7 \\
			Belagiannis\&Zisserman FG'17 \cite{belagiannis2016recurrent}, &95.2 &89.0 &81.5 &77.0 &83.7 &87.0 &82.8 &85.2 \\
			Yu et al. ECCV'16 \cite{yu2016deep},  &87.2 &88.2 &82.4 &76.3 &91.4 &85.8 &78.7 &84.3 \\
			Rafi et al. BMVC'16 \cite{rafi2016efficient},  &95.8 &86.2 &79.3 &75.0 &86.6 &83.8 &79.8 &83.8 \\
			Yang et al. CVPR'16 \cite{yang2016end}, &90.6 &78.1 &73.8 &68.8 &74.8 &69.9 &58.9 &73.6 \\
			Chen\&Yuille NIPS'14 \cite{chen2014articulated}, &91.8 &78.2 &71.8 &65.5 &73.3 &70.2 &63.4 &73.4 \\
			Fan et al. CVPR'15 \cite{fan2015combining}, &92.4 &75.2 &65.3 &64.0 &76.7 &68.3 &70.4 &73.0 \\
			Tompson et al. NIPS'14 \cite{tompson2014joint}, &90.6 &79.2 &67.9 &63.4 &69.5 &71.0 &64.2 &72.3 \\
			Pishchulin et al. ICCV'13 \cite{pishchulin2013strong}, &87.2 &56.7 &46.7 &38.0 &61.0 &57.5 &52.7 &57.1 \\
			Wang\&Li et al. CVPR'13 \cite{fwang2013pose}, &84.7 &57.1 &43.7 &36.7 &56.7 &52.4 &50.8 &54.6 \\
			\hline
		\end{tabular}
	\end{center}
	
	\caption{Comparisons of PCK@0.2 score on the LSP test set.}
	\label{table:LSP-results}
\end{table*}

\paragraph{PCP}
Table \ref{table:LSP-results-PCP} reports the PCP at threshold of $0.5$.
\begin{table*}[t]
	\footnotesize
	\tabcolsep=0.2cm
	\begin{center}
		\begin{tabular}{l c c c c c c c c}
			\hline 
			Method & Torso & U.Leg & L.Leg & U.Arm & Forearm & Head & Total \\ [0.2ex]
			\hline
			Ours & 98.6  & 95.8  & 93.6  & 90.7  & 84.2  & 96.4 & \textbf{92.3}   \\
			\hline
			
			Bulat\&Tzimiropoulos. ECCV'16 \cite{bulat2016human},  &97.7 	&92.4 	&89.3 	&86.7 	&79.7 	&95.2 	&88.9 \\
			Wei et al. CVPR'16 \cite{wei2016convolutional}, &98.0 	&82.2 	&89.1 	&85.8 	&77.9  &95.0 &88.3 \\
			Insafutdinov et al. ECCV'16 \cite{insafutdinov2016deepercut},  &97.0 	&90.6 	&86.9 	&86.1 	&79.5 	&95.4 	&87.8 \\
			Yu et al. ECCV'16 \cite{yu2016deep},  &98.0 	&93.1 	&88.1 	&82.9 	&72.6 	&83.0 	&85.4 \\
			Pishchulin et al. CVPR'16 \cite{pishchulin2016deepcut},  &97.0 	&88.8 	&82.0 	&82.4 	&71.8 	&95.8 	&84.3 \\
			Lifshitz et al. ECCV'16 \cite{lifshitz2016human}, &97.3 	&88.8 	&84.4 &80.6 	&71.4 	&94.8 	&84.3 \\
			Belagiannis\&Zisserman FG'17 \cite{belagiannis2016recurrent}, &96.0 	&86.7 	&82.2 	&79.4 	&69.4 	&89.4 	&82.1 \\
			Rafi et al. BMVC'16 \cite{rafi2016efficient},  &97.6 	&87.3 	&80.2 	&76.8 	&66.2 	&93.3 	&81.2 \\
			Yang et al. CVPR'16 \cite{yang2016end}, &95.6 	&78.5 	&71.8 	&72.2 	&61.8 	&83.9 	&74.8 \\
			Chen\&Yuille NIPS'14 \cite{chen2014articulated}, &96.0 	&77.2 	&72.2 	&69.7 	&58.1 	&85.6 	&73.6 \\
			Fan et al. CVPR'15 \cite{fan2015combining}, &95.4 	&77.7 	&69.8 	&62.8 	&49.1 	&86.6 	&70.1 \\
			Tompson et al. NIPS'14 \cite{tompson2014joint}, &90.3 	&70.4 	&61.1 	&63.0 	&51.2 	&83.7 	&66.6 \\
			Pishchulin et al. ICCV'13 \cite{pishchulin2013strong}, &88.7 	&63.6 	&58.4 	&46.0 	&35.2 	&85.1 	&58.0 \\
			Wang\&Li et al. CVPR'13\cite{fwang2013pose}, &87.5 	&56.0 	&55.8 	&43.1 	&32.1 	&79.1 	&54.1 \\
			\hline
		\end{tabular}
	\end{center}
	\caption{Comparisons of PCP@0.5 score on the LSP test set.}
	\label{table:LSP-results-PCP}
\end{table*}

\begin{figure*}[h]
	\begin{center}
		\includegraphics[width=0.9\linewidth]{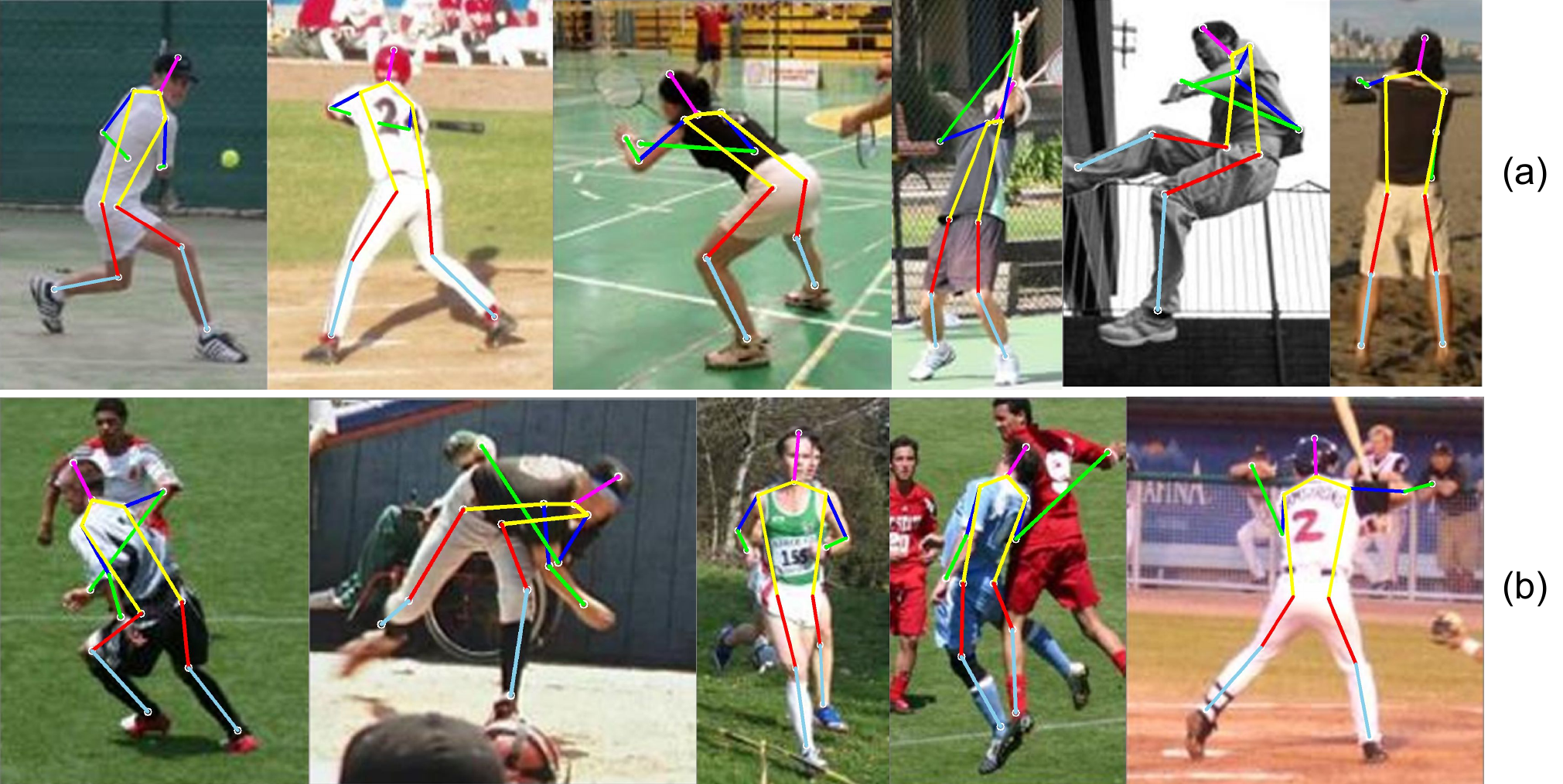}
	\end{center}
	\caption{Failure cases on LSP dataset: (a) Ambiguity caused by full occlusion of 2 or more adjacent body parts; (b) Regression mistake caused by the concurrence of body part noise from other persons and full occulusion of less than 2 body parts. }
	\label{fig:fail-cases}
\end{figure*}

\subsection{Algorithm Performance Analysis and Ablation Study}
\label{sec:component-analysis}

Since the ground truth of MPII dataset is not publicly available and it is forbidden to frequently submit MPII test results to the official, we perform component analysis of our proposed method on the LSP dataset.
We analyze the contribution of each component in Table \ref{table:component-analysis}. 

We compare the proposed inception-resnet module and the basic resnet module employed by stacked hourglass networks \cite{newell2016stacked}. 
Since their performance is not reported on LSP dataset, we implement their network within our system to render fair comparisons.
Under identical settings, our network with inception-resnet module achieves superior performance over that with basic resnet module by improving the accuracy by $1.1$\%. 
We also compare our network under standard training with the same network under knowledge projection and guided learning. Results show that better performance is achieved with knowledge guided training with an accuracy improvement of $1.3$\%.
We then analyze contributions of other techniques employed mainly during testing, i.e., flipping the image, testing the image at multiple scales, and using proposed NMS algorithm for pose estimation. 
Testing on original and flipped images improves performance by 0.7\%, while testing on both original and $0.75$ scales further improves performance by another $0.7$\%. Cross-heatmap non-maximum suppression improves the PCK value by $0.5$\%.

It should be noted that our implementation\footnote{Code and models available at: http://github.com/Guanghan/GNet-pose}  in \textit{PyCaffe} \cite{jia2014caffe} may not fully reproduce equivalent performance on MPII dataset of the hourglass network \cite{newell2016stacked}, which is implemented in \textit{Torch} \cite{torch}. 
However, we discuss with performance analysis to show that our proposed knowledge guided training is able to improve the performance on top of existing deep neural network. We expect that the same performance gain can be achieved on other network structures.


\section{Conclusion}
\label{sec:conclusions}
In this work, we have proposed to encode and inject external human knowledge into deep neural networks to guide its training process with learned projections for more effective human pose estimation. 
We adopt the stacked hourglass design and propose to use inception-resnet as the building block of our fractal network to regress human pose into heatmaps with no explicit graphical modeling.
Utilizing a multi-resolution feature representation with guided learning, the network learns an empirical set of low and high-level features which are typically more tolerant to variations in the training set.
Knowledge-guided learning is a generic scheme that can be potentially used to aid other deep neural network training tasks.
The effectiveness of the proposed inception-resnet module and the benefit in guided learning with knowledge projection is evaluated on two widely used benchmarks. 


%





\ifCLASSOPTIONcaptionsoff
  \newpage
\fi



%


\bibliographystyle{IEEEtran}
\bibliography{ning}

%








\end{document}